\documentclass[letterpaper]{article} 
\usepackage{aaai2026}  
\usepackage{times}  
\usepackage{helvet}  
\usepackage{courier}  
\usepackage[hyphens]{url}  
\usepackage{graphicx} 
\usepackage{natbib}  
\usepackage{caption} 
\usepackage{algorithm}
\usepackage{algorithmic}
\usepackage{booktabs}
\usepackage{amsmath}
\usepackage{multirow}
\usepackage{siunitx}
\usepackage{newfloat}
\usepackage{listings}
\DeclareCaptionStyle{ruled}{labelfont=normalfont,labelsep=colon,strut=off} 
\floatstyle{ruled}
\newfloat{listing}{tb}{lst}{}
\floatname{listing}{Listing}
\nocopyright 

\title{Gender Bias in Emotion Recognition by Large Language Models}
\author{
    Maureen Herbert\equalcontrib,
    Katie Sun\equalcontrib,
    Angelica Lim,
    Yasaman Etesam
}
\affiliations{
    \textsuperscript{}Simon Fraser University\\

    Burnaby, BC, Canada\\
    mrh9, ksa138, angelica, yetesam@sfu.ca
}

\usepackage{bibentry}

\begin{document}

\maketitle

\begin{abstract}
The rapid advancement of large language models (LLMs) and their growing integration into daily life underscore the importance of evaluating and ensuring their fairness. In this work, we examine fairness within the domain of emotional theory of mind, investigating whether LLMs exhibit gender biases when presented with a description of a person and their environment and asked, ``How does this person feel?''. Furthermore, we propose and evaluate several debiasing strategies, demonstrating that achieving meaningful reductions in bias requires training based interventions rather than relying solely on inference-time prompt-based approaches such as prompt engineering, etc.
\end{abstract}


\section{Introduction}

As artificial agents increasingly interact with humans, it is essential for them to possess emotional intelligence~\cite{bera2019emotionally} and be able to perceive and infer human emotions reliably. However, emotion recognition is inherently subjective and our interpretations of others’ feelings are shaped by both societal norms and individual perspectives~\cite{etesam2025vision}. \citet{plaza2024angry} showed that such biases may also emerge in LLMs when asking LLMs for an emotion label given a situation and a gender. In this paper, we examine these biases using context-rich image descriptions and a multi-label setup, focusing specifically on how gendered perceptions influence the interpretation of emotional expressions. In contrast to \citet{plaza2024angry}, we ask the model to identify the other person’s emotion rather than describing how the LLM itself would feel in the situation. We formulate this as a multi-label classification task covering a broad range of 26 emotions derived from the EMOTIC dataset~\cite{kosti2019context}.
Gender bias is often shown in subtle ways that reinforce systemic inequalities. A study by \citet{condry1976sex} illustrated this effect experimentally: when participants observed identical emotional responses from infants, they tended to describe the behavior as ``anger'' when the infant was labeled boy, and as ``fear'' when labeled girl. This finding highlights how observers project gendered stereotypes onto emotional expressions. These biases in humans are reflected in the data that models are trained on, consequently causing the models to learn and reproduce those biases.

Large language models (LLMs), trained on vast datasets of human-generated text, may internalize perceptual biases. Our research investigates how this inheritance manifests in an emotion recognition task and reproduces the gender biases observed in humans. Specifically, we use the NarraCap captions~\cite{etesam2024contextual} constructed for the EMOTIC~\cite{kosti2019context} dataset as descriptions of context-rich scenarios where people experience different emotions. We input the same caption to LLMs but with different genders and observe whether the models’ predictions change accordingly. By analyzing predicted emotion labels across gender modified captions, we examine to what extent different LLMs such as GPT, Mistral~\cite{jiang2023mistral}, and LLaMA~\cite{touvron2023llama} contain these subconscious biases that influence the model's perception when analyzing the captions. 

\begin{figure}[t!]
    \centering
    \includegraphics[width=0.45\textwidth]{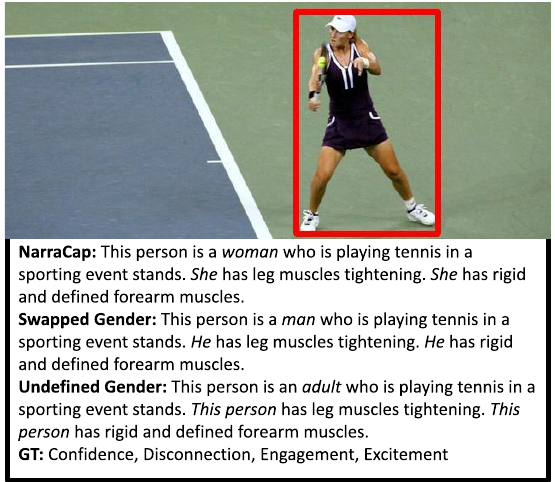} 
    \caption{An EMOTIC image~\cite{kosti2019context} with the corresponding NarraCap caption, along with swapped and undefined gender versions. GT represents the ground truth emotion labels chosen by annotators.}
    \label{fig:narracap}
\end{figure}

In this work, we relied on data augmentation to achieve debiasing. Specifically, we randomly sampled captions from NarraCap~\cite{etesam2024contextual} with EMOTIC~\cite{kosti2019context} ground truth emotion labels collected from annotators and then expanded them by: 1) swapping the gender and 2) removing the gender from the caption, while retaining the original ground truth emotion labels for all versions (see Fig.~\ref{fig:narracap}). By fine-tuning the model on this augmented dataset, we aimed to desensitize it to gender.
To summarize, our contributions are:
\begin{itemize}
    \item Proposing an investigation of gender biases in LLMs within a multi-label emotion estimation framework.
    \item Evaluating gender biases in the emotion recognition task across various LLMs, including GPT-4, GPT-5, Mistral, and LLaMA.
    \item Investigating different debiasing approaches to reduce gender influence in LLMs’ emotion predictions, including both inference time prompt-based and training-based methods.  
\end{itemize}

\section{Related Work}

\subsection{Context-based Emotion Recognition}
In this paper, we investigate gender bias in LLMs in the context of emotion recognition, which aims to understand a person’s apparent emotions.
We emphasize that the person's true emotions cannot be reliably inferred, as ground truth is often unavailable; instead, this task focuses on predicting the emotion labels provided by annotators, i.e., ``apparent emotion recognition''~\cite{etesam2025vision}.
Early work largely focused on facial emotion recognition \cite{pantic2000expert, ko2018brief, mellouk2020facial, chu2016selective}, but Barrett et al. (2019) highlighted key challenges, arguing that facial movements alone do not reliably indicate emotion categories across situational contexts in daily life~\cite{barrett2019emotional}. To incorporate contextual cues, the EMOTIC dataset \cite{kosti2019context} was introduced as a multi label dataset containing diverse images of people in varying situations experiencing different emotions, alongside a two-branch CNN baseline model to solve the task. Building on this, several approaches have employed attention mechanisms to improve estimation \cite{le2022global, mittel2023peri}, while others have explored multi-branch architectures for distinct context interpretations \cite{mittal2020emoticon, yang2022emotion}. More recent methods leverage image captioning to transform visual content into text and extract co-occurrence relationships between words~\cite{chen2023incorporating, de_Lima_Costa_2023_CVPR}.
Recent advances in large language models have introduced their use in contextual emotion recognition, either by first generating captions and feeding them to an LLM ~\cite{yang2023contextual, etesam2024contextual}, or by directly employing multimodal large language models ~\cite{etesam2023emotional, etesam2024contextual, zhang2024vision}.

\subsection{Emotional Intelligence and LLMs}
\citet{mayer2011emotional} describes emotional intelligence (EI) as the ``ability to monitor one's own and others' feelings and emotions, to discriminate among them and to use this information to guide one's thinking and actions.''~\cite{mayer2011emotional}. 
Language plays a crucial role in emotion perception and reasoning \cite{lindquist2015role, lieberman2007putting, lindquist2006language, lindquist2013s, gendron2012emotion}, and LLMs have shown emerging capacities in these domains. 
Early work demonstrated latent capabilities for reasoning in large language models (LLMs) \cite{huang2023towards}, including some sub-tasks on emotion inference \cite{mao2022biases, sap2022neural}.
Recent studies have begun systematically assessing LLMs’ EI. Psychometric evaluations found above-average EQ scores but notable variation across models \cite{wang2023emotional}. 
EMOBENCH \cite{sabour2024emobench} addressed benchmark limitations by testing emotional understanding and application, revealing substantial human–model gaps.
Other work emphasized the need for non-deterministic assessments more aligned with human EI \cite{dalal2025llms}. By contrast, \citet{schlegel2025large} reported that frontier models (e.g., GPT-4, Claude, Gemini) outperformed humans on five EI tests and could even generate novel test items.
Overall, LLMs display promising yet uneven emotional reasoning abilities, with systematic evaluation of their alignment to human EI still an open challenge.


\subsection{Social Bias in LLMs}
``Social bias broadly encompasses disparate treatment or outcomes between social groups that arise from historical and structural power asymmetries''~\cite{gallegos2024bias}. Prior studies reveal varying degrees of such biases in large language models (LLMs)~\cite{bai2023fairmonitor, zhao2023gptbias}.
Social biases have also been studied in specific domains such as auto-generated code~\cite{ling2025bias} recommendation~\cite{tommasel2024fairness, li2025can}, ranking~\cite{wang2024large}, political ideology~\cite{lin2024investigating}, and gender stereotypes~\cite{dwivedi2023breaking, dong2024disclosure, sorokovikova2025surface}.
\citet{plaza2024angry} also show that societal biases and stereotypical patterns appear in emotion attribution across LLMs.
Recent work further highlights that while LLMs may appear unbiased under explicit bias benchmarks, they can still harbor implicit biases that remain hidden without more nuanced evaluation~\cite{bai2025explicitly}. 
To address these challenges, a variety of bias mitigation techniques have been proposed~\cite{gallegos2024bias}. Specific strategies include prompt engineering and in-context learning~\cite{dwivedi2023breaking, chhikara2024few}, hyperparameter tuning, instruction guiding, and debias tuning~\cite{dong2024disclosure}, as well as model fine-tuning approaches~\cite{lin2024investigating}.

\section{Methodology}

\begin{table*}[t!]
\centering
\small
\begin{tabular}{@{}lcccccccccccc@{}}
\toprule
 & \multicolumn{2}{c}{GPT4o-mini} & \multicolumn{2}{c}{GPT5-mini} & \multicolumn{2}{c}{Mistral-instruct} & \multicolumn{2}{c}{TinyLLaMA} & \multicolumn{2}{c}{DeepSeek} & \multicolumn{2}{c}{LLaMA}\\ 
\cmidrule(lr){2-3} \cmidrule(lr){4-5} \cmidrule(lr){6-7} \cmidrule(lr){8-9} \cmidrule(lr){10-11} \cmidrule(lr){12-13}
 & chi2 & p & chi2 & p & chi2 & p & chi2 & p & chi2 & p & chi2 & p\\ 
\midrule
\textbf{suffering} & 3.28&0.07	& 0.01&0.92&0.58&0.45&0.52&0.47 & 0.04 &0.85&0.28&0.60\\
\textbf{pain} & 0.00 & 1.00	& 0.00	& 1.00 & 0.31 & 0.58&0.17&0.68 & 0.21 &0.65&0.00	& 0.99\\
\textbf{sadness} & 0.69	& 0.41 & 0.00 & 0.95 & 0.07 &	0.79 & 0.01 & 0.94 & 0.41 &0.52&1.24&0.27\\
\textbf{aversion} & 0.36 & 0.55 & 0.12 & 0.73 & 1.76& 0.19	& 0.01 & 0.91 &0.00 & 1.00&0.36&0.55\\
\textbf{disapproval} & 0.00	& 1.00 & 0.00 & 1.00 & 0.00	& 1.00 & 0.48 &0.49 & 0.01 & 0.94&0.61&0.44\\
\textbf{anger} & 0.78 & 0.38 & 0.00 & 0.97 & 0.13 & 0.72 & 0.28 & 0.60 &  0.01 & 0.93&0.76&0.38\\
\textbf{fear} & 0.05 & 0.82 & 0.00 & 0.99 & 0.06 & 0.80 & 0.06 & 0.80 & 0.50 & 0.48&0.04&0.84\\
\textbf{annoyance} & 0.06 & 0.81 & 0.00 & 1.00 & 0.12 & 0.73 & 0.41 & 0.52 & 0.15 & 0.70&0.13&0.72\\
\textbf{fatigue} & 1.42 & 0.23 & 0.05 & 0.82 & 1.16 & 0.28 & 0.13 & 0.72 &  0.00 & 0.97&1.17&0.28\\
\textbf{disquietment} & 1.04 & 0.31 & 0.00 & 1.00 & 0.50 & 0.48 & 0.00 & 1.00 & 0.75& 0.39 &0.95&0.33\\
\textbf{doubt/confusion} & \textbf{7.49} & \textbf{0.01} & 0.03 & 0.87 & 0.00 & 1.00 & 0.00 & 1.00 & 0.12 & 0.73&0.00&0.96\\
\textbf{embarrassment} & 0.00 & 1.00 & 0.40 & 0.53 & 0.02 & 0.89 & 0.41 & 0.52 &  0.04 & 0.85&1.29&0.26\\
\textbf{disconnection} & 0.39 & 0.53 & 0.61 & 0.44 & 2.12 & 0.15 & 1.93 & 0.16 & 0.00 & 0.99&0.10&0.75\\
\textbf{affection} & 1.86 & 0.17 & 2.41 & 0.12 & 0.03 & 0.87 & 0.00 & 1.00 & 0.73 & 0.39&0.05&0.83\\
\textbf{confidence} & 0.25 & 0.62 & 0.79 & 0.37 & 0.45 & 0.50 & 3.56 & 0.06& 0.13 & 0.72&2.28&0.13\\
\textbf{engagement} & 0.11 & 0.74 & 0.01 & 0.94 & 0.94 & 0.33 & 0.94 & 0.33 & 0.35 & 0.55&0.19&0.67\\
\textbf{happiness} & 0.31 & 0.58 & 0.35 & 0.56 & 2.29& 0.13 & 0.02 & 0.87 & 1.14 & 0.29&0.02&0.89\\
\textbf{peace} & 0.02 & 0.90 & 0.04 & 0.84 & 0.96 & 0.33 & 0.82 & 0.36 & 0.21 & 0.65&1.90&0.17\\
\textbf{pleasure} & 0.00 & 1.00 & 0.16 & 0.69 & \textbf{7.28} & \textbf{0.01} & 0.00 & 0.99 & 0.01  & 0.93 &0.11&0.74\\
\textbf{esteem} & 0.50 & 0.48 & 0.25 & 0.62 & 1.35 & 0.25 & 0.35 & 0.56 & 1.87 & 0.17&0.02&0.88\\
\textbf{excitement} & 0.00 & 0.99 & 0.01 & 0.91 & 0.01 & 0.94 & 0.00 & 0.98 &0.046 & 0.83&0.45&0.50\\
\textbf{anticipation} & 0.01 & 0.92 & 0.60 & 0.44 & 0.27 & 0.60 & 0.01 & 0.91 & 0.24  &0.62&\textbf{8.36}&\textbf{0.00}\\
\textbf{yearning} & 0.01 & 0.91 & 0.36 & 0.55 & 0.00 & 1.00 & 0.36 & 0.55 & 0.11 & 0.74&0.07&0.80\\
\textbf{sensitivity} & 0.97 & 0.33 & 0.00 & 0.97 & 1.17 & 0.28 & 0.76 & 0.38 & -  &-&\textbf{7.09}&\textbf{0.01}\\
\textbf{surprise} & 0.06 & 0.81 & 0.00 & 1.00 & 0.18 & 0.67 & 0.43 & 0.51 & 0.11 & 0.73&0.21&0.65\\
\textbf{sympathy} & 0.20 & 0.65 & 0.09 & 0.77 & 1.66 & 0.20 & 0.42 & 0.52 & 0.00&  0.96&1.04&0.31\\

\bottomrule
\end{tabular}
\caption{We employed multiple LLMs to perform the emotion recognition task. In this table, we report the Chi-square test values for the association between man and woman variables.}
\label{tab:llms}
\end{table*}

Gender bias refers to a preference for or prejudice against one gender over another~\cite{sun2019mitigating}. In this paper, we investigate gender biases in the emotion recognition task in LLMs. Specifically, we examine whether large language models (LLMs) generate consistent outputs when image captions are modified to reflect different genders. 
\subsection{Defining and Measuring Gender Bias}
In this work, we adopt an equal distribution baseline as our definition of an unbiased model. Specifically, a model is considered gender-unbiased if it predicts each emotion label equally for men and women—that is, maintaining a 50:50 distribution across genders for every emotion.
This definition acts as a practical reference point for several reasons:
\begin{itemize}
    \item It acknowledges that there is no objective ``ground truth'' distribution of emotions by gender that could serve as an alternative reference.
    \item It gives us a clear, consistent, quantifiable baseline to measure the bias magnitudes across diverse categories. 
\end{itemize}

We emphasize that this 50:50 baseline is a measurement framework rather than a claim about human emotional expression. While human observers do exhibit gender bias in emotion perception (\citet{condry1976sex}), our goal is to quantify and compare biases in LLMs using a consistent, neutral benchmark. In Sec. 3.3.1, we also explore how the model would respond if training data were not 50:50.

\subsection{Dataset}
To examine the effect of gender on emotion estimation in LLMs, we follow the ``captioning $+$ LLM'' methodology proposed by \citet{etesam2024contextual}. This method converts an image of a person into a textual description (NarraCap in Fig.~\ref{fig:narracap}) and subsequently performs emotion inference on that text description. We utilize the NarraCap captions~\cite{etesam2024contextual}, which are generated by passing EMOTIC~\cite{kosti2019context} images through CLIP~\cite{radford2021learning} to answer the questions ``who'', ``what'', ``where'', and ``how''. EMOTIC contains context-rich images of people experiencing different emotions, with multi-label ground-truth annotations covering 26 categories, which were selected by clustering 400 affect-related words using the ‘visual separability’ criterion~\cite{kosti2019context}. This emphasis on context in EMOTIC makes the generated NarraCap captions rich in contextual information. 
The answer to the ``who'' question in NarraCap provides the gender and age of the person, considering only man and woman genders. While it has been shown that these captions can be improved and there is a need for better captions for more accurate emotion estimation~\cite{yang2023contextual, etesam2023emotional}, it is important to note that the focus of this work is not on the emotion recognition task itself, but on gender biases in emotion recognition. Consequently, in this study, we do not compare the generated emotion labels with ground truth labels, but rather with the generated labels corresponding to the caption with opposite gender.

For our study, we randomly selected $1000$ samples from the NarraCap captions corresponding to the EMOTIC validation set and expanded them by 1) swapping the gender (e.g., changing \textit{boy} to \textit{girl}, \textit{man} to \textit{woman}, and \textit{he} to \textit{she}) and 2) neutralizing it (e.g., using \textit{adult} instead of \textit{man} or \textit{woman}, and \textit{this person} instead of \textit{he} or \textit{she}). We want to emphasize that, although the original image distribution may contain an underlying bias toward women or men, this does not affect the proposed approach as we assume a 50–50 distribution. You can see an example in Fig.~\ref{fig:narracap}, which shows the original NarraCap caption, the swapped gender version, and the undefined caption, while the ground truth labels remain the same across all three versions. Although the ground truth labels are not used for evaluation, they are required for fine-tuning.

\subsection{Evaluating LLMs}

\begin{figure*}[ht!]
    \centering
    \includegraphics[width=1\textwidth]{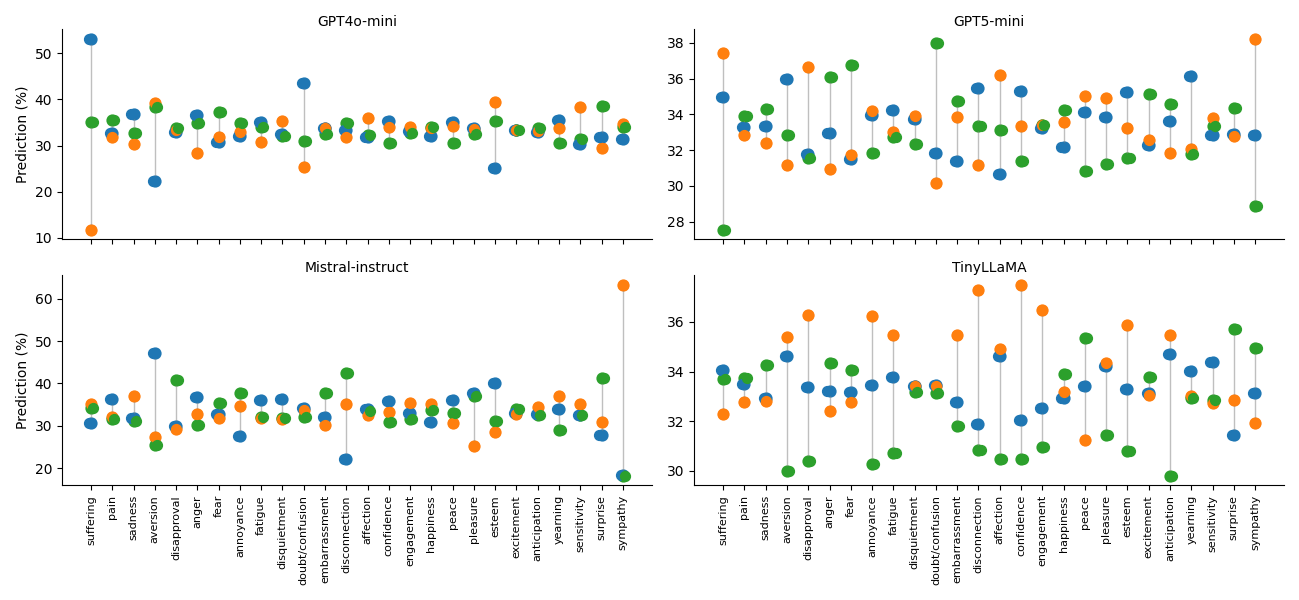} 
    \caption{This figure shows the frequency of emotion labels predicted by GPT-4, GPT-5, Mistral, and TinyLLaMA for captions with man (blue), woman (orange), and undefined (green) genders. To better illustrate the differences across genders, the predictions were normalized based on each emotion label.}
    \label{fig:mfn}
\end{figure*}

On our curated dataset, we compare different LLMs by examining the distributions of predicted emotion labels for captions with woman, man, and undefined subjects. A model without gender bias should treat these captions similarly, producing comparable distributions across genders.
We assessed the following models on these three sets: Mistral, TinyLLaMA, GPT-4o-mini, GPT-5-mini, DeepSeek, and LLaMA. This evaluation is performed in a zero-shot setting, without any task-specific training. The same prompt is used for all models:

\begin{quote}
\texttt{From this list of emotions: \{EMOTIC 26 labels\} pick the most likely emotions this person feels simultaneously. \\Return ONLY comma-separated emotions. No explanations. \\Caption: \{NarraCap caption\} \\Emotions:}
\end{quote}

\subsubsection{Simulating Non-50:50 Emotion Distributions}
We acknowledge that the distribution of emotional expressions with respect to gender may differ  depending on cultural context. To evaluate the impact of source training data on model bias, we simulate two extreme possible scenarios: one where the subject of all captions are women, and another where all caption subjects are men. We first randomly selected $200$ captions where not only the subject of the caption is a woman/man, but also the ground truth from the EMOTIC dataset is woman/man. This is to ensure consistency between the gender identified by the annotators and the emotion labels they chose for the person. We then fine-tuned the LLM using caption–emotion label pairs using only woman as the subject (FT-W) or a man for the subject (FT-M).

\subsection{Debiasing LLMs}
We then used the open-source Mistral-instruct-7B~\cite{jiang2023mistral} model and applied different techniques to reduce gender bias. These techniques include  inference time prompt based approaches, such as prompt engineering, in-context learning, and chain-of-thought reasoning, as well as a fine-tuning method.

\subsubsection{Prompt Engineering}
To encourage the model to generate emotion labels without considering the person's gender, we add \textit{``Disregard any gender bias you have.''} to the prompt.

\subsubsection{In-context Learning}
We also experimented with enhancing the prompt by adding examples to our original prompt. To assist with debiasing, we include two similar captions in which the only difference is the person’s gender, while the expected emotion labels remain unchanged. This indicates that changing the gender should not influence the predicted labels:

\begin{quote}
\texttt{<original prompt> $+$ \\Example:\\
Caption: \\The woman wiped her eyes and smiled softly as she looked at the photo.\\ Emotions: Sadness, Happiness, Peace, Yearning, Sensitivity, Engagement\\ \\ Example:\\ Caption: The man wiped his eyes and smiled softly as he looked at the photo.\\Emotions: Sadness, Happiness, Peace, Yearning, Sensitivity, Engagement}
\end{quote}

\subsubsection{Chain-of-Thought (CoT)}
\citet{etesam2023emotional} showed that the chain-of-thought technique can help large vision-language models infer people's emotions more accurately. Here, we explore whether this technique can also help LLMs be less emotionally biased. To this end, we use this prompt:
\begin{quote}
\texttt{From this list of emotions: \{EMOTIC 26 labels\} pick the most likely emotions this person feels simultaneously.\\ Explain the reasoning behind your choice(s) and then give the emotion label(s).\\Example:\\Caption: "The woman wiped her eyes and smiled softly as she looked at the photo."\\Reasoning and emotion labels: She feels the pain of missing someone: Sadness. She wishes she could be with the person or relive the moment: Yearning. She smiles softly, recalling a joyful memory: Happiness, Peace. The photo evokes deep emotional response: Sensitivity. She is fully absorbed in the memory: Engagement.\\Caption: \{NarraCap caption\}\\ Reasoning and emotion labels:}
\end{quote}

\subsubsection{Fine-tuning using Caption-Emotion Labels pairs (FT)}
We also explored fine-tuning the model using 
LoRA~\cite{hu2021lora}. For this task, we randomly selected $100$ samples from the NarraCap captions in the validation set that were not included in our $1000$ test samples. As shown by \cite{etesam2024contextual}, only $100$ samples are sufficient to fine-tune these models.
Following the same procedure used to create our test data, we expanded these $100$ samples into $200$ caption–emotion label pairs by changing the gender in the captions while keeping the same set of emotion labels. We augmented the data by multiplying each caption by $10$ and randomly shuffling the emotion labels each time to eliminate any effects of order.
We then fine-tuned Mistral on this dataset, enabling the model to learn that similar captions with different genders should yield the same set of emotion labels. The prompt we used during fine-tuning is similar to our original prompt.

\section{Experiments}

\begin{table}[ht]
\small
\centering
\begin{tabular}{@{}l|lccc@{}
  l 
  S[table-format=2.2(3)] 
  S[table-format=2.2(3)] 
  S[table-format=2.2(3)] 
  S[table-format=2.2(3)] 
  @{}}

\toprule
& {Model} & {Female} & {Male} & {Undefined} \\
\midrule
\multirow{6}{*}{\rotatebox[origin=c]{90}{LLMs}}
& GPT-4o mini & 4737 & 4734 & 4817 \\
& GPT-5 mini & 4610 & 4553 & 4575 \\
& DeepSeek & 3820 & 3759 & 3861 \\
& TinyLLaMA & 12356 & 12531 & 11961 \\
& LLaMA & 5869 & 5843 & 5851 \\
& Mistral Instruct & 3008 & 2920 & 2978\\

\midrule
\multirow{4}{*}{\rotatebox[origin=c]{90}{De-biased}}
& Prompt eng & 3192 & 3120& 3178 \\
& In-context & 5204 & 5106 & 5191 \\
& CoT & 8711 & 8488 & 8830 \\
& FT & 5739 & 5711 & 5677 \\
\bottomrule


\end{tabular}
\caption{Number of predicted labels for each model and gender.}

\label{tab:freq}
\end{table}

\begin{table*}[t]
\centering
\small
\begin{tabular}{@{}lcccccccccccc@{}}
\toprule
 & \multicolumn{2}{c}{Zero-shot} & \multicolumn{2}{c}{Prompt-eng} & \multicolumn{2}{c}{In-context} & \multicolumn{2}{c}{CoT} & \multicolumn{2}{c}{FT} \\ 
\cmidrule(lr){2-3} \cmidrule(lr){4-5} \cmidrule(lr){6-7} \cmidrule(lr){8-9} \cmidrule(lr){10-11} 
 & chi2 & p & chi2 & p & chi2 & p & chi2 & p & chi2 & p \\ 
\midrule
\textbf{suffering} & 0.58 & 0.45 & 0.28 & 0.60 & 0.01 & 0.92 & 0.00 & 1.00 & 0.00 & 1.00\\
\textbf{pain} & 0.31 & 0.58 & 0.26 & 0.61 & 0.00 & 1.00 & 0.06 & 0.80 & 0.00 & 1.00 \\
\textbf{sadness} & 0.07 & 0.79 & 0.67 & 0.41 & 2.41 & 0.12 & 0.02 & 0.88 & 0.00 & 1.00 \\
\textbf{aversion} & 1.76 & 0.19 & 0.00 & 1.00 & \textbf{4.29} & \textbf{0.04} & 1.93 & 0.16 & 0.00 & 1.00 \\
\textbf{disapproval} & 0.00 & 1.00 & 0.03 & 0.86 & 0.28 & 0.60 & 0.02 & 0.90 & 0.11 & 0.75 \\
\textbf{anger} & 0.13 & 0.72 & 0.01 & 0.92 & 0.00 & 1.00 & 0.59 & 0.44 & 0.00 & 1.00 \\
\textbf{fear} & 0.06 & 0.80 & 0.15 & 0.69 & 0.07 & 0.80 & 0.35 & 0.55 & 0.05 & 0.83 \\
\textbf{annoyance} & 0.12 & 0.73 & 0.18 & 0.67 & 0.12 & 0.73 & 0.10 & 0.75 & 0.16 & 0.69 \\
\textbf{fatigue} & 1.16 & 0.28 & 2.07 & 0.15 & \textbf{4.88} & \textbf{0.03} & 0.46 & 0.50 & 0.00 & 1.00\\
\textbf{disquietment} & 0.50 & 0.48 & 0.07 & 0.79 & 1.59 & 0.21 & 0.55 & 0.46 & 0.01 & 0.91 \\
\textbf{doubt/confusion} & 0.00 & 1.00 & 0.01 & 0.94 & 0.19 & 0.66 & 1.57 & 0.21 & 0.05 & 0.82 \\
\textbf{embarrassment} & 0.02 & 0.89 & 0.61 & 0.43 & 0.16 & 0.68 & 1.01 & 0.31 & 0.00 & 1.00 \\
\textbf{disconnection} & 2.12 & 0.15 & 0.02 & 0.88 & 2.00 & 0.16 & 1.12 & 0.29 & 0.62 & 0.43 \\
\textbf{affection} & 0.03 & 0.87 & 0.25 & 0.62 & 0.00 & 1.00 & 1.47 & 0.23 & 1.67 & 0.20\\
\textbf{confidence} & 0.45 & 0.50 & 0.97 & 0.33 & 2.99 & 0.08 & 0.03 & 0.85 & 0.01 & 0.92 \\
\textbf{engagement} & 0.94 & 0.33 & 0.36 & 0.55 & 0.25 & 0.61 & 0.14 & 0.71 & 0.01 & 0.92\\
\textbf{happiness} & 2.29 & 0.13 & 2.15 & 0.14 & \textbf{6.47} &\textbf{ 0.01} & \textbf{5.77} & \textbf{0.02} & 0.01 & 0.93 \\
\textbf{peace} & 0.96 & 0.33 & 0.61 & 0.43 & 0.00 & 1.00 & 0.26 & 0.61 & 1.68 & 0.19 \\
\textbf{pleasure} & \textbf{7.28} & \textbf{0.01} & \textbf{3.88 }& \textbf{0.05} & 2.92& 0.09 & 0.44 & 0.51 & 0.25 & 0.61 \\
\textbf{esteem} & 1.35 & 0.25 & 0.84 & 0.36 & \textbf{6.21} & \textbf{0.01} & 0.42 & 0.52 & 0.79 & 0.38 \\
\textbf{excitement} & 0.01 & 0.94 & 0.00 & 0.98 & 1.33 & 0.25 & 0.13 & 0.72 & 0.03 & 0.86 \\
\textbf{anticipation} & 0.27 & 0.60 & 0.09 & 0.77 & 1.33 & 0.25 & 1.32 & 0.25 & 0.00 & 0.94 \\
\textbf{yearning} & 0.00 & 1.00 & 0.31 & 0.58 & 0.00 & 1.00 & 0.39 & 0.53 & 0.18 & 0.67\\
\textbf{sensitivity} & 1.17 & 0.28 & 3.64 & 0.06 & \textbf{5.33} & \textbf{0.02} & 3.19 & 0.07 & 1.08 & 0.30 \\
\textbf{surprise} & 0.18 & 0.67 & 0.00 & 1.00 & 0.00 & 1.00 & 0.02 & 0.88 & 0.00 & 0.99 \\
\textbf{sympathy} & 1.66 & 0.20 & 1.69 & 0.19& 2.10 & 0.15 & 1.21 & 0.27 & 1.43 & 0.23 \\

\bottomrule
\end{tabular}
\caption{We applied different debiasing techniques, using Mistral Instruct-7B as the base model for all methods. In this table, we report the results of a Chi-square test to determine whether there is a statistically significant (p<0.05) association between man and woman variables.}
\label{tab:debias}
\end{table*}

We passed the captions to different LLMs and collected the frequency of each predicted emotion label. It is important to note that some models generated emotion labels outside the 26 predefined categories (e.g., exhaustion). We excluded those labels and only considered predictions that matched the 26 EMOTIC emotion labels.
The experiments were conducted using a single NVIDIA GeForce RTX 3090 GPU.

\subsection{Evaluation Metric}
For the evaluation metric, we either report the number of times an emotion is predicted for different genders, normalized for each emotion (Fig.~\ref{fig:mfn}), or use the Chi-square ($\chi^2$) test. The Chi-square test is a statistical method used to determine whether there is a significant association between categorical variables (in this case, man and woman). It compares the observed frequencies in each category to the expected frequencies if there were no relationship between the variables. The $\chi^2$ value (Chi-square statistic) measures how much the observed data deviate from the expected values, the larger it is, the greater the difference. The $p$-value indicates the probability that such a difference could occur by chance; a small $p$-value suggests that the observed association is statistically significant, meaning it is unlikely to have occurred randomly.
Some models failed to predict certain labels entirely, for example, DeepSeek did not predict ``sensitivity'' for either man or woman (see Table~\ref{tab:llms}).
In such cases, it is not possible to compute ($\chi^2$).

\subsection{LLMs}
For LLMs, we configured \texttt{do\_sample=False} and \texttt{max\_new\_tokens=64}. When employing chain-of-thought prompting, however, we set \texttt{max\_new\_tokens=256}.

\subsubsection{Mistral} We used Mistral-7B-Instruct-v0.3\footnote{https://huggingface.co/mistralai/Mistral-7B-Instruct-v0.3}, an LLM which is an instruction tuned version of Mistral-7B-v0.3.

\subsubsection{TinyLLaMA} We used TinyLlama-1.1B\footnote{https://huggingface.co/TinyLlama/TinyLlama-1.1B-Chat-v1.0}, a pretrained LLaMA model with 1.1 billion parameters trained on 3 trillion tokens. The model was trained over a 90-day period using 16 NVIDIA A100-40GB GPUs.

\subsubsection{LLaMA} We used llama-3.3-70b-versatile\footnote{\url{https://console.groq.com/docs/model/llama-3.3-70b-versatile}}, a model based on Meta's LLaMA 3.3 and fine-tuned for helpfulness and safety.

\subsubsection{GPT-4} We used GPT-4o mini\footnote{https://platform.openai.com/docs/models/gpt-4o-mini} which is a smaller, faster, and cheaper version of GPT-4o that handles text and image inputs. This is the only multi modal model included in our experiments.

\subsubsection{GPT-5} We utilized GPT-5 mini\footnote{https://platform.openai.com/docs/models/gpt-5-mini} which is a smaller, faster, and cheaper version of GPT-5.

\subsubsection{Deepseek} We utilized deepseek-chat\footnote{https://docs.deepseekapi.io/deepseek-api/chat/} which is a Chat completion model.
\subsection{Fine-tuning}
For our fine-tuning approaches, we used LoRA and \texttt{Mistral-7B-Instruct-v0.3} as the base model. The LoRA parameters were set as follows: \(r=8\), \(\text{lora\_alpha}=16\), and target modules are \texttt{q\_proj, k\_proj, v\_proj, and lm\_head}.

\section{Results}
\begin{table}[t!]
\centering
\small
\begin{tabular}{@{}lcccc@{}}
\toprule
 & FT-W & FT-M & chi & p \\ 
\midrule

\textbf{suffering} & 79& 20& 30.27& \textbf{0.00}\\
\textbf{pain} & 51 & 17	& 13.85	& \textbf{0.00} \\
\textbf{sadness} & 48	& 33& 1.54 & 0.21 \\
\textbf{aversion} & 38 & 37 & 0.01& 0.94 \\
\textbf{disapproval} & 84	& 60 & 2.25 & 0.13  \\
\textbf{anger} & 20 & 17 & 0.01 & 0.91 \\
\textbf{fear} & 23 & 59 & 17.66 & \textbf{0.00} \\
\textbf{annoyance} & 51 & 60 & 1.30 & 0.26 \\
\textbf{fatigue} & 71 & 24 & 19.30 & \textbf{0.00}  \\
\textbf{disquietment} & 76 & 188 & 56.10 & \textbf{0.00}\\
\textbf{doubt/confusion} & 76 & 25 & 21.53 & \textbf{0.00} \\
\textbf{embarrassment} & 37 & 28 & 0.50 & 0.48 \\
\textbf{disconnection} & 254 & 150 & 20.13 & \textbf{0.00}\\
\textbf{affection} & 201 & 136 & 8.27 & \textbf{0.00}  \\
\textbf{confidence} & 639 & 653 & 2.97 & 0.08  \\
\textbf{engagement} & 937 & 995 & 9.59 & \textbf{0.00}  \\
\textbf{happiness} & 827 & 825 & 2.14 & 0.14 \\
\textbf{peace} & 188 & 158 & 0.83 & 0.36 \\
\textbf{pleasure} & 386 & 376 & 0.36 & 0.55\\
\textbf{esteem} & 224 & 142 & 13.03 & \textbf{0.00} \\
\textbf{excitement} & 678 & 719 & 6.44 & \textbf{0.01} \\
\textbf{anticipation} & 953 & 965 & 3.82 & \textbf{0.05}\\
\textbf{yearning} & 56 & 38 & 2.01 & 0.16\\
\textbf{sensitivity} & 22 & 45 & 8.92 & \textbf{0.00} \\
\textbf{surprise} & 102 & 36 & 26.50 & \textbf{0.00} \\
\textbf{sympathy} & 544 & 403 & 12.93 & \textbf{0.00} \\
\midrule
Sum  & 6665 & 6209 & - & -\\
\bottomrule
\end{tabular}
\caption{Results for 1000 test captions with gender removed (e.g. "This person is an adult who..."). Emotions are inferred by fine-tuned models trained on woman caption-emotion pairs (FT-W) or man caption-emotion pairs (FT-M). The bolded values show significant (p<0.05) differences between FT-W and FT-M output counts.}
\label{tab:ftwm}
\end{table}

We evaluated the frequency of emotion labels predicted by multiple LLMs for captions with man, woman, and undefined genders. Fig.~\ref{fig:mfn} visualizes these distributions for 4 of these models (GPT-4o mini, GPT-5 mini, Mistral-instruct, TinyLLaMA), normalized per emotion label to highlight relative differences across gender variants.
We observe deviations in the predicted emotion distributions across man, woman, and undefined captions.
It is important to note that the total number of predictions per model and for each gender varies across different models (see Table~\ref{tab:freq}). Specifically, among the various LLMs, TinyLLaMA tends to predict more labels across all settings, and chain-of-thought model also produce a higher number of predictions across de-biased models. Furthermore, except for TinyLLaMA, the other models tend to predict fewer labels for captions referring to men compared to those referring to women or undefined genders.

We applied Chi-square tests to examine whether there was a statistically significant association between predicted emotion labels and gender (man vs. woman) across different models (Table~\ref{tab:llms}). Among these models, GPT5-mini, TinyLLaMA and DeepSeek show no significant gender bias ($p \ge 0.05$), whereas GPT4o-mini, Mistral instruct, and LLaMA exhibit bias for at least one emotion.
We observe that different models exhibit varying degrees of bias toward specific emotions, which may be attributed to differences in the data on which they were trained. 
Table~\ref{tab:debias} reports the $\chi^2$ statistics and corresponding p-values for prompting and fine-tuning methods using Mistral Instruct-7B as the base model. Certain methods exhibited significant deterioration; for example, the in-context learning technique introduces significant biases across different emotion labels. Notably, \emph{aversion}, \emph{fatigue}, \emph{happiness}, \emph{esteem}, and \emph{sensitivity} were significant under in-context prompting ($p \le 0.05$); however, the significance of \emph{pleasure} was reduced compared to original Mistral Instruct.
The other prompt-based mitigation techniques, while not as detrimental as in-context learning, still did not improve the bias, which is in line with the findings of \citet{kuan2025gender}. However, fine-tuned model (FT) eliminated detectable bias, with all emotions yielding non-significant associations. These results suggest that, while prompting methods do not mitigate gender-related patterns in emotion estimation, fine-tuning reduces potential gender bias across the evaluated emotional categories to some extent.
\subsection{Non-50:50 Distribution}
In Table~\ref{tab:ftwm}, we present the results for 1000 test-set captions with gender information removed, evaluated using models fine-tuned on caption–emotion label pairs containing only man or only woman subjects. Although the inputs to both fine-tuned models are identical, we observe markedly different distributions of predicted emotions. We compute the chi-square statistic and p-value, where the
null hypothesis is that fine-tuning in either direction would result in equivalent shifts. We notice significant differences (p<0.05) toward women for suffering, pain, fatigue, doubt/confusion, disconnection, affection, esteem, surprise and sympathy. For men, we observe significant differences (p<0.05) for fear, disquietment, engagement, and anticipation. One possible interpretation is that fine-tuning with 200 randomly selected samples of non-50:50 data can result in unevenness across emotions represented, and data balancing across all emotions may be necessary. Another possibility is that the underlying model contains bias that is difficult to shift with small-scale fine-tuning of an arbitrary gender distribution. Future work can investigate this area of cultural adaptation. Another observation from this table is that the FT-M and FT-W models might steer predictions toward emotion distributions associated with a single gender. While inputting captions with different gender indicators such as ``man'', ``woman'', or ``undefined'' to each model may yield similar prediction distributions and appear unbiased, this can be problematic in a settings where emotion distributions across genders should not be similar. In such cases, the model may learn and consequently produce output distributions biased toward the dominant gender (i.e., the one more prevalent in the training data). Therefore, while the FT-W model may be reliable for female subjects and the FT-M model for male subjects, neither model can be assumed reliable for individuals outside this binary gender categorization.




\section{Conclusion}
In this paper, we evaluated gender biases in LLMs on the apparent emotion recognition task. We observed that most of the models exhibit significant gender biases for at least one emotion label. We also proposed different techniques to mitigate these biases, including inference-time prompting and fine-tuning. Our results show that while inference-time prompting did not improve the biases, fine-tuning techniques can be an effective way to mitigate them.

\section{Limitations}
Although our results demonstrate measurable gender-related emotional biases in LLMs, it is important to consider several limitations when interpreting them.

Our study uses text captions from static visual scenes, which only partially capture real-world emotional complexity. In natural interactions, emotion perception depends on tone, body language, context, and interpersonal dynamics, so the biases we observe may not fully generalize to multi modal or conversational settings.
 
Different LLMs tend to output varying numbers of emotion labels per caption. Since EMOTIC allows multiple concurrent emotions, this variability can influence the normalized frequency distributions and $\chi^2$ statistics, potentially masking bias-related patterns.

We restrict our analysis to the 26 EMOTIC emotion categories and a limited set of models (GPT-4, GPT-5, Mistral, LLaMA, TinyLLaMA, DeepSeek). Expanding the study to include other models such as Claude or Gemini,  or 
alternative emotional taxonomies (e.g., Plutchik’s wheel) 
could reveal different trends.
 
The NarraCap dataset and our augmentation procedure include only binary gender categories (man/woman), so non-binary and gender-diverse identities are not represented. This reflects dataset limitations rather than theoretical intent, and future work should aim to cover all gender diversities.
 
It is worth considering that emotional expression may vary across genders. In such a scenario, an LLM’s gender-based probabilistic estimates of emotion could be justified. However, we do not have data to either confirm or deny this. 
Similarly, cultural factors may influence how emotions are expressed and perceived across different genders, leading to differences in emotion distribution patterns.


Overall, these limitations suggest that the reported biases and mitigation outcomes should be interpreted as indicative rather than conclusive. Future research should explore multi modal evaluations, more diverse gender representations, larger training corpora, and more comprehensive emotional frameworks to better understand and mitigate bias in emotional theory of mind within LLMs.



\section{Ethical Impact Statement}
This research explores gender biases in LLMs for the apparent emotion recognition task. In this section, we discuss the ethical implications of this work.

\textbf{Potential Negative Societal Impact.} 
LLM-powered software can unintentionally spread and reinforce stereotypes, including gender biases, when such biases exist in the models. As these tools are put to more widespread use, they may influence public perception and behavior in subtle but significant ways. While the LLMs examined in this study improve accessibility, they also allow for widespread deployment without adequate safeguards. This could unintentionally amplify latent biases on a large scale, potentially shaping users’ perceptions, decisions, and interactions even through applications that appear neutral or unbiased. The authors do not support applying this research in ways that perpetuate harmful stereotypes or deepen gender biases.

\textbf{Limits of Generalizability.} 
The gender bias identified in this study is specific to the pre-trained models we evaluated, which include both open-source models and proprietary models accessed via API. Each model was trained on its own dataset and likely reflects societal biases present in that data. Our study focuses on detecting these biases in specific contexts and may not capture the full range of cultural variations in emotional expression, particularly across non-Western cultures and marginalized communities.

\textbf{Other Issues} This work relies on pre-trained LLMs, including Mistral-7B-Instruct-v0.3, TinyLlama-1.1B, llama-3.3-70b-versatile, GPT-4o mini, and GPT-5 mini. Additionally, we conducted lightweight fine-tuning on the pre-trained Mistral-7B-Instruct-v0.3 model. 
We recognize that the large-scale computational resources used for pre-training LLMs contribute to a tangible environmental footprint. It is important to acknowledge this impact responsibly.


\textbf{Potential Negative Societal Impact.} 
LLM-powered software can unintentionally spread and reinforce stereotypes, including gender biases, when such biases exist in the models. As these tools are put to more widespread use, they may influence public perception and behavior in subtle but significant ways. While the LLMs examined in this study improve accessibility, they also allow for widespread deployment without adequate safeguards. This could unintentionally amplify latent biases on a large scale, potentially shaping users’ perceptions, decisions, and interactions even through applications that appear neutral or unbiased. The authors do not support applying this research in ways that perpetuate harmful stereotypes or deepen gender biases.

\textbf{Limits of Generalizability.} 
The gender bias identified in this study is specific to the pre-trained models we evaluated, which include both open-source models and proprietary models accessed via API. Each model was trained on its own dataset and likely reflects societal biases present in that data. Our study focuses on detecting these biases in specific contexts and may not capture the full range of cultural variations in emotional expression, particularly across non-Western cultures and marginalized communities.

\textbf{Other Issues} This work relies on pre-trained LLMs, including Mistral-7B-Instruct-v0.3, TinyLlama-1.1B, llama-3.3-70b-versatile, GPT-4o mini, and GPT-5 mini. Additionally, we conducted lightweight fine-tuning on the pre-trained Mistral-7B-Instruct-v0.3 model. 
We recognize that the large-scale computational resources used for pre-training LLMs contribute to a tangible environmental footprint. It is important to acknowledge this impact responsibly.

\section{Acknowledgments}
This research was supported by the Canada CIFAR AI Chairs Program and the Rajan Scholar Research Fund.

\bibliography{aaai2026}

\newpage

\end{document}